\documentclass[runningheads]{llncs}
\usepackage[utf8]{inputenc} 
\usepackage[T1]{fontenc}    
\usepackage[colorlinks, pagebackref=true, citecolor=blue, urlcolor=blue, linkcolor=blue]{hyperref}
\usepackage{url}            
\usepackage{booktabs}       
\usepackage{amsfonts}       
\usepackage{nicefrac}       
\usepackage{microtype}      
\usepackage{xcolor}         
\usepackage[ruled]{algorithm2e}
\usepackage{caption}
\usepackage{subcaption}
\usepackage{graphicx}

\usepackage[numbers,sort, sort&compress]{natbib}

\SetKwComment{Comment}{$\triangleright$\ }{}

\usepackage{amsmath,amsfonts,bm}

\def\eqref#1{equation~\ref{#1}}

\def\1{\bm{1}}

\def\rmI{{\mathbf{I}}}

\def\vc{{\bm{c}}}

\def\vm{{\bm{m}}}

\def\vr{{\bm{r}}}

\def\vw{{\bm{w}}}
\def\vx{{\bm{x}}}
\def\vy{{\bm{y}}}
\def\vz{{\bm{z}}}

\def\mC{{\bm{C}}}

\def\mI{{\bm{I}}}

\def\mQ{{\bm{Q}}}
\def\mR{{\bm{R}}}

\def\mW{{\bm{W}}}
\def\mX{{\bm{X}}}
\def\mY{{\bm{Y}}}

\DeclareMathAlphabet{\mathsfit}{\encodingdefault}{\sfdefault}{m}{sl}
\SetMathAlphabet{\mathsfit}{bold}{\encodingdefault}{\sfdefault}{bx}{n}
\newcommand{\tens}[1]{\bm{\mathsfit{#1}}}

\def\tY{{\tens{Y}}}

\def\sG{{\mathbb{G}}}

\newcommand{\R}{\mathbb{R}}

\newcommand{\spm}[1]{{\scriptstyle \pm {#1}}}

\newcommand{\CE}{\operatorname{CE}}
\newcommand{\MLP}{\operatorname{MLP}}
\newcommand{\pcom}{\Pi^{\operatorname{CoM}}}

\begin{document}
\title{MiDi: Mixed Graph and 3D Denoising Diffusion for Molecule Generation}
\titlerunning{MiDi}
%
\author{Clément Vignac\inst{1}\textsuperscript{*}\and
Nagham Osman\inst{2}\textsuperscript{*} \and
Laura Toni\inst{2} \and
Pascal Frossard\inst{1}}
\authorrunning{Vignac, Osman et al.}
%
\institute{LTS4, EPFL, Lausanne, Switzerland \and 
EEE Dept, University College London, London, United Kingdom\\
}
\maketitle              
\begin{abstract}
  This work introduces MiDi, a novel diffusion model for jointly generating molecular graphs and their corresponding 3D arrangement of atoms. 
  Unlike existing methods that rely on predefined rules to determine molecular bonds based on the 3D conformation, MiDi offers an end-to-end differentiable approach that streamlines the molecule generation process. Our experimental results demonstrate the effectiveness of this approach. On the challenging GEOM-DRUGS dataset, MiDi generates 92\% of stable molecules, against $6\%$ for the previous EDM model that uses interatomic distances for bond prediction, and $40\%$ using EDM followed by an algorithm that directly optimize bond orders for validity.
  Our code is available at \url{github.com/cvignac/MiDi}.

\end{abstract}
\section{Introduction}

\begin{figure}[b]
\vspace{-0.3cm}
    \centering
    \includegraphics[width=\textwidth]{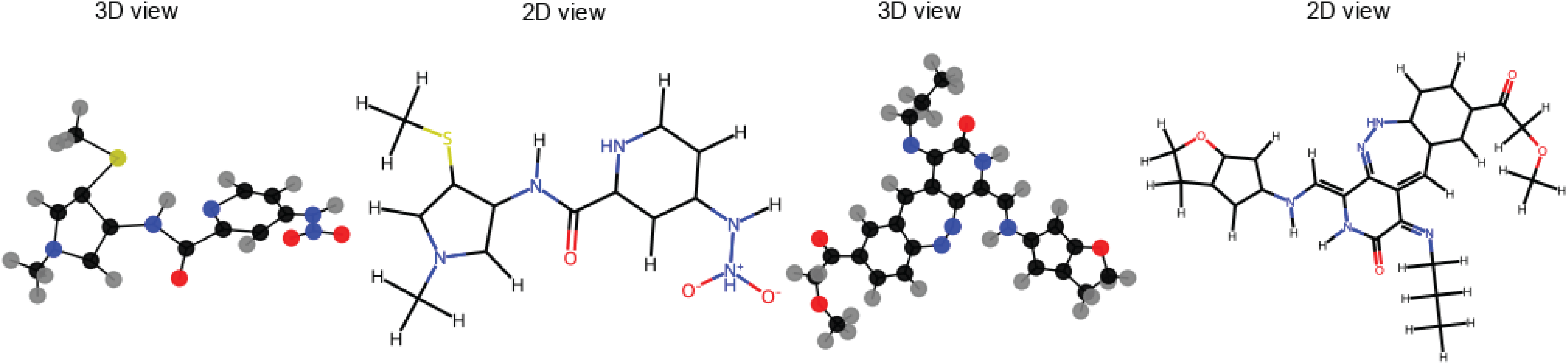}
    \caption{Samples from our model. MiDi generates simultaneously a 2D graph structure and a 3D conformation that is consistent with this structure.}
    \label{fig:my_label}
\end{figure}

Modern drug discovery requires the development of effective machine learning models that can correctly capture the vast chemical space and sample from it. These models need to understand properties of molecules that depend both on their molecular graph and their conformation in the 3D space. 
The molecular graph (or 2D structure) determines the existence and type of the chemical bonds, and allows the identification of functional groups in a compound. This provides information about its chemical properties and enables to predict synthetic pathways. 
On the other hand, the 3D conformation of a compound plays a key role in its interaction with other molecules, and governs in particular its biological activity and binding affinity to proteins.
To explore the chemical space adequately, it is therefore crucial to consider both aspects simultaneously.

Unfortunately, existing generative models for molecules are restricted to one of these two data modalities. While models that exclusively generate molecular graphs have been vastly researched \citep{du2022molgensurvey}, 
current 3D molecule generation are on the contrary only trained to generate conformers, thus ignoring bond information. These models rely on a subsequent step that predicts the 2D structure using either interatomic distances \cite{satorras2021en_flows, hoogeboom2022equivariant} or chemical software such as OpenBabel \cite{gebauer2019symmetry}.  
As a result, these models are not end-to-end differentiable, which hampers their ability to be fully optimized for various downstream tasks. This severely limits the potential of 3D molecule generators, particularly for complex tasks like pocket-conditioned generation.

We propose here a new model, called \underline{Mi}xed Graph+3D Denoising \underline{Di}ffusion (MiDi), which overcomes this limitation by simultaneously generating a molecular graph and its corresponding 3D coordinates. MiDi represents molecules as graphs embedded in 3D that contain node features (atom types and formal charges) and edges features (bond types).
Our model progressively corrupts data with noise, and trains a neural network to predict clean data from noisy inputs. New molecules can then be generated by sampling pure noise and iteratively denoising it with the neural network, similarly to other diffusion models \citep{sohldickstein2015diffusion, ho2020denoising}.
As the model is trained to denoise both the graph and 3D coordinates in tandem, it is able to produce stable molecular graphs that are consistent with the generated conformers.

While previous diffusion models for molecules relied on either Gaussian noise or discrete diffusion, MiDi uses both noise models simultaneously: the 3D coordinates are corrupted with Gaussian noise, while the other components use discrete diffusion which was found to be effective for graph generation \cite{vignac2022digress, haefeli2022diffusion}. To further enhance the quality of the generated samples, we introduce a noise schedule whose parameters are adjusted to each component. Specifically, we add noise to the atom types and formal charges at a faster rate than to the coordinates and bond types. This encourages the denoising network to first focus on generating a realistic 3D conformation and corresponding bond types, before refining the atom types and formal charges.  

Our second contribution considers the denoising network: the Transformer architecture we propose incorporates a novel \emph{rEGNN} layer, which improves upon the popular EGNN layers \cite{satorras2021egnn} by leveraging features that are not translation-invariant. We show that, due to the use of Gaussian noise in the zero center-of-mass subspace of the molecules, the resulting model is nevertheless equivariant to translations and rotations, which is crucial for achieving high performance.

We showcase the effectiveness of our model on unconditional molecule generation. On the challenging GEOM-DRUGS dataset, the previous EDM model \citep{hoogeboom2022equivariant} can generate stable molecules at a rate of $5.5\%$, which can be improved to $40.3\%$ by using the Open Babel bond prediction algorithm \citep{o2011open}. In contrast, the proposed method achieves $91.6\%$ of stable molecules, demonstrating the superiority of our end-to-end differentiable approach. Furthermore, MiDi can be readily applied to various drug-discovery tasks beyond unconditional generation, which confirms its versatility and potential for improving drug discovery pipelines. 

\section{Related Work}

\subsubsection{Concurrent work}
Concurrently to our work, \cite{hua2023mudiff} and \cite{peng2023moldiff} also proposed 2D+3D diffusion models for molecule generation. These models also leverage Gaussian diffusion for the 3D coordinates and discrete diffusion for the graph, using absorbing transitions in \cite{peng2023moldiff} and uniform transitions in \cite{hua2023mudiff}. Each model also features unique contributions: \cite{hua2023mudiff} proposes richer positional encodings for the transformer layer, while \cite{peng2023moldiff} introduces a guidance mechanism to help the network predict accurate bond lengths. MiDi is the only model that  improves upon the standard EGNN layers, and that is capable of handling formal charges. It is also the only model that presents results on the more complex GEOM-DRUGS dataset with explicit hydrogens.

\subsubsection{Molecule generation in 3D} The idea of representing molecules as attributed 3D point cloud has been used both in one-shot settings \cite{satorras2021en_flows} and in autoregressive methods such as GSchNet\cite{gebauer2019symmetry} . Recently, the Equivariant Diffusion Model (EDM) \cite{hoogeboom2022equivariant} was proposed for this task, improving significantly over previous results. This model was later extended by limiting the message-passing computations to neighboring nodes \cite{huang2022mdm} and using a more expressive denoising network \citep{morehead2023geometry}.
All these diffusion models can be conditioned on molecule-level properties using guidance mechanisms \cite{bao2022equivariant} or on another point cloud. Conditioning on a second point cloud has been employed to generate molecules that bind to a specific protein \cite{corso2022diffdock, schneuing2022structure} and to generate linkers between molecular fragments \cite{igashov2022equivariant}. 
The main drawback of these models is that they do not learn the connectivity structure of the molecule. It needs to be obtained in a second stage using interatomic distances \cite{satorras2021en_flows, hoogeboom2022equivariant} or specialized software such as Open Babel \cite{o2011open}. This results in limited performance for complex molecules, but also prevents end-to-end differentiability for downstream applications.

\subsubsection{Graph Generation}

Another line of work has focused on generating graphs without associated 3D coordinates. Early denoising diffusion models for this task used Gaussian noise applied to the adjacency matrix \cite{jo2022score,huang2022graphgdp} or graph eigenvalues \cite{luo2022fast}. \cite{vignac2022digress} and \cite{haefeli2022diffusion} however found that discrete diffusion is more effective, as it better respects the discrete nature of graphs. These diffusion models tend to outperform autoregressive methods except on validity metrics, as autoregressive models can perform validity checks at each sampling step \cite{liu2018constrained, liao2019efficient, mercado2021graph}.
In contrast to the proposed method, which operates at the node level, fragment-based methods \cite{hajduk2007decade, jin2020hierarchical, maziarz2021learning} learn to combine chemically-relevant substructures from a fixed or learned dictionary \cite{wang2022retrieval}, but are harder to adapt to 3D. 

In order to generate molecules in 3D, graph-based models could be combined with conformer generation models \cite{xu2022geodiff, corso2022diffdock} which predict a 3D structure from an input graph. As these methods assume that the graph is known, they are able to exploit symmetries of the molecule (such as rotatable bonds), which is more difficult on unconditional generation tasks. Unfortunately, combining graph generation and conformer generation models would again break end-to-end differentiability and restrict performance.

\subsubsection{Protein Generation}

While existing diffusion models for molecules operate on molecules of moderate size (up to 180 atoms), recent diffusion models for proteins have managed to scale to much larger structures \cite{watson2022broadly, ingraham2022illuminating, wu2022protein,shi2022protein}. These methods leverage the chain structure of proteins, which implies that the adjacency matrix does not need to be predicted. Furthermore, instead of predicting 3D coordinates for each atom, they only predict the angles between successive $C_\alpha$ carbons, which significantly reduces the degrees of freedom and encodes roto-translation invariance in the representation. Those improvements are unfortunately specific to chain graphs, and cannot be used for arbitrary molecules.

\section{Background} \label{sec:background}

\subsubsection{Denoising Diffusion Models} \label{subsec:diffusion}

\begin{table}[t]
    \centering  \caption{Gaussian and categorical distributions enable the efficient computation of the key quantities involved in training diffusion models and sampling from them. Formulas for all parameters can be found in Section \ref{subsec:diffusion}.}
    \begin{tabular}{l l c} \toprule
   Noise model & $~$Gaussian diffusion & Discrete diffusion \\ \midrule
       $q(\vz_t | \vz_{t-1})$ & $~\mathcal N(\alpha_t \vz_{t-1},~ \sigma_t^2 \mI) $ &  $\vz_{t-1}~ \mQ_{t}$ \\
       $q(\vz_t | \vx)$  & $~\mathcal N(\bar \alpha_t \vx,~ \bar \sigma_t^2 \mI) $ & $\vx~ \bar \mQ_t $\\
       $\int_{x}  p_\theta(\vz_{t-1}| \vx, \vz_t)dp_\theta(\vx|\vz_t)$  & $~\mathcal N(\mu_t \hat \vx + \nu_t \vz_t, \tilde \sigma_{t}^2 \mI)$ & $~\propto \sum_{x} p_\theta(x) (\vz_t \mQ_t' \odot \vx \bar \mQ_{t-1})$ \\
    \bottomrule
    \end{tabular}
    
    \label{tab:noise-models}
\end{table}

Diffusion models consist of two essential elements: a noise model and a denoising neural network. The noise model $q$ takes as input a data point $x$ and generates a trajectory of increasingly corrupted data points $(z_1, ..., z_T)$. The corruption process is chosen to be Markovian, i.e., 
$$q(z_1, \dots, z_T | x) = q(z_1 | x) \prod_{t=2}^T q(z_t | z_{t-1}).$$

The denoising network $\phi_\theta$ takes noisy data $z_t$ as input, and learns to invert the diffusion trajectories. While it would be natural to naively train the network to predict $z_{t-1}$, this strategy would lead to noisy targets, as $z_{t-1}$ depends on the sampled diffusion trajectory. Instead, modern diffusion models \citep{sohldickstein2015diffusion, song2019estimatinggradients, ho2020denoising} predict the clean input $x$ from $z_t$, or equivalently, the noise added to it. 
The diffusion sequences are then inverted by marginalizing over the network predictions $p_\theta(x|z_t)$:
\begin{equation} \label{eq:marginalization}
p_\theta(z_{t-1} | z_t) = \int_{x}  p_\theta(z_{t-1}~|~ x, z_t) ~ dp_\theta(x|z_t)
\end{equation}

Although Eq. \ref{eq:marginalization} leads to more efficient training, it requires the efficient computation of $p_\theta(z_{t-1} | x, z_t)$ and the integral, which is not always possible. Two main frameworks have been proposed under which Eq. \ref{eq:marginalization} is tractable: Gaussian noise, which is suitable for continuous data, and discrete state-space diffusion for categorical data. Table \ref{tab:noise-models} summarizes the main properties of the two related noise models.

Gaussian diffusion processes are defined by $q(z_t | z_{t-1}) \sim \mathcal N(\alpha_t \vz_t, \sigma_t^2 \mI)$, where $(\alpha_t)_{t\leq T}$ controls how much signal is retained at each step and $(\sigma_t)_{t \leq T}$ how much noise is added. As normal distributions are stable under composition, we have $q(z_t | x) \sim \mathcal N(\bar \alpha_t \vz_t, \bar \sigma_t^2 \mI)$, with $\bar \alpha_t = \prod_{s=1}^t \alpha_s$ and $\bar \sigma_t^2 = \sigma_t^2 - \alpha_t^2$. While any noise schedule is in principle possible, variance-preserving processes are most often used, which satisfy $\bar \alpha_t^2 + \bar \sigma_t^2 = 1$.
The posterior of the transitions conditioned on $x$ can also be computed in closed-form. It satisfies
$$    q(z_{t-1} | z_t, x) \sim \mathcal N (\mu_t ~\vx + \nu_t ~\vz_t, \tilde\sigma_t^2 \rmI),$$
with $\mu_t = \bar\alpha_s (1- \alpha_t^2 \bar\sigma_{t-1}^2 / \bar \sigma_t^2)$, $\nu_t =\alpha_t \bar\sigma_{t-1}^2 / \bar \sigma_t^2  $ and $\tilde \sigma_t = \bar \sigma_{t-1}^2 (1 - \alpha_t^2  \bar \sigma_{t-1}^2 /  \bar\sigma_t^2)$. \\

On the contrary, discrete diffusion considers that data points $x$ belong to one of $d$ classes \cite{austin2021structured}.
The transition matrices $(\mQ_1, ..., \mQ_T)$ are square matrices of size $d \times d$ that represent the probability of jumping from one class to another at each time step. Given previous state $z_{t-1}$, the noise model for the next state $z_t$ is a categorical distribution over the $d$ possible classes which reads as $q(z_{t} |z_{t-1}) \sim \mathcal C(\vz_{t-1} \mQ_{t})$, where $\vz_{t-1}$ is a row vector encoding the class of $z_{t-1}$. 
Since the process is Markovian, we simply have $q(z_t=j | x) = [x \bar \mQ^t]_{j}$ with $ \bar \mQ^t= \mQ^1 \mQ^2... \mQ^t$.
The posterior distribution $q(z_{t-1} | z_t, x)$ can also be computed in closed form using Bayes rule and the Markovian property. If $\odot$ denotes a pointwise product and $\mQ'$ is the transpose of $\mQ$, it can be written as
$$q(z^{t-1} | z^t, x) \propto \vz^t ~(\mQ^t)' \odot \vx ~\bar \mQ^{t-1}.$$

\subsubsection{SE(3)-Equivariance with Diffusion Models}
Molecules are dynamic entities that can undergo translation and rotation, and the arrangement of their atoms does not have a predetermined order. To effectively model molecules using generative models and avoid augmenting the data with random transformations, it is essential to ensure that the models are equivariant to these inherent symmetries.
In diffusion models, equivariance to a transformation group $\sG$ can be achieved through several conditions. First, the noise model must be equivariant to the action of $\sG$: $\forall g \in \sG, q(g. z_t | g.x) = q(z_t | x)$. Second, the prior distribution $q_\infty$ used at inference should be invariant to the group action, i.e., $q_\infty(g. z_T) = q_\infty(z_T)$, and this noise should be processed by an equivariant neural network in order to ensure that $p_\theta(g. z_{t-1} | g. z_t) = p_\theta(z_{t-1} | z_t)$. Finally, the network should be trained with a loss function that satisfies $l(p_\theta(g. x | g . z_t), g. x) = l(p_\theta(x|z_t), x)$. Together, these requirements create an architecture that is agnostic to the group elements used to represent the training data \cite{xu2022geodiff, hoogeboom2022equivariant, vignac2022digress}.

To ensure equivariance to the special Euclidean group SE(3), a number of architectures have been proposed as possible denoising networks for a diffusion model \cite{thomas2018tensorfield, brandstetter2021geometric, gasteiger2021gemnet, liao2022equiformer}. However, these networks can be computationally expensive due to their manipulation of spherical harmonics. As a result, many generative models for molecules \cite{hoogeboom2021argmax, igashov2022equivariant, schneuing2022structure, huang2022mdm} use the more affordable EGNN layers \cite{satorras2021egnn}. At a high level, EGNN recursively updates the coordinates $(\vr_i)$ of a graph with node features $(\vx_i)$ and edge features $(\vy_{ij})$ using:
$$ \vr_i \gets \vr_i + \sum_{j} c_{ij}~ m(||\vr_i - \vr_j||, \vx_i, \vx_j, \vy_{ij}) (\vr_j - \vr_i) $$ 
The crucial feature of this parameterization is that the message function $m$ takes only rotation-invariant arguments. This, combined with the linear term in $\vr_j - \vr_i$, ensures that the network is rotation-equivariant. Finally, we note that the normalization term $c_{ij} = ||\vr_i - \vr_j|| + 1$ is necessary for numerical stability when concatenating many layers.

\section{Proposed Model}
We now present the Mixed Graph+3D denoising diffusion (MiDi) model. We represent each molecule as a graph $G=(\vx, \vc, \mR, \mY)$, where $\vx$ and $\vc$ are vectors of length $n$ containing the type and formal charge associated to all atoms. The $n \times 3$ matrix $\mR=[\vr_i]_{1\leq i\leq n}$ contains the coordinates of each atom, and $\mY$ is an $n \times n$ matrix containing the bond types. Similarly to previous diffusion models for graphs, we consider the absence of a bond as a particular bond type and generate dense adjacency tensors. We denote the one-hot encoding of $\vx$, $\vc$, and $\mY$ by $\mX$, $\mC$, and $\tY$, respectively. Time steps are denoted by superscripts, so, for example, $\vr_i^t$ denotes the coordinates of node $v_i$ at time $t$. The transpose of matrix $\mX$ is denoted by $\mX'$.

\subsection{Noise Model}

Our noise model corrupts the features of each node and edge independently, using a noise model that depends on the data type. For the positions, we use a Gaussian noise within the zero center-of-mass (CoM) subspace of the molecule $\bm\epsilon \sim \mathcal N^{\text{CoM}}(\alpha^t \mR^{t-1}, ~(\sigma^t)^2 \mI)$, which is required to obtain a roto-translation equivariant architecture \cite{xu2022geodiff}. This means that the noise follows a Gaussian distribution on the linear subspace of dimension $3 (n-1)$ that satisfies $\sum_{i=1}^n \bm\epsilon_i = 0$. 

For atom types, formal charges and bond types, we use discrete diffusion, where the noise model is a sequence of categorical distributions. We choose the marginal transition model proposed in \cite{vignac2022digress}. For instance, when $\vm \in \R^{a}$ represents the marginal distribution of atom types in the training set, we define $\mQ_x^t = \alpha^t \mI + \beta^t ~\bm 1_a \vm'$. We similarly define $\mQ_c^t$ and $\mQ_y^t$. The resulting noise model is given by:
$$ q(G^t | G^{t-1}) \sim \mathcal N^{\text{CoM}}(\alpha^t \mR^{t-1}, ~(\sigma^t)^2 \mI) \times \mathcal C(\mX^{t-1} \mQ_x^t) \times \mathcal C( \mC^{t-1} \mQ_c^t )\times \mathcal C(\tY^{t-1} \mQ_y^t).$$ 

When generating new samples, we define the posterior as a product as well:
\begin{equation*}
p_\theta(G^{t-1} | G^t) = \prod_{1 \leq i \leq n} p_\theta(\vr_i^{t-1} | G^t) p_\theta(\vx_i^{t-1} | G^t) p_\theta(\vc_i^{t-1} | G^t) \prod_{1\leq i, j \leq n} p_\theta(\mY_{ij}^{t-1} | G^t),
\end{equation*}
We calculate each term by marginalizing over the network predictions. For instance,
\begin{align*}
    p_\theta(x_i^{t-1} | G^t) &= \int_{\vx_i}  p_\theta(x_i^{t-1}~|~ x_i, G^t) ~ dp_\theta(x_i|G^t) \\
    &= \sum_{x \in \mathcal X}  q(x_i^{t-1}|x_i=x, G^t) ~ p_\theta^X(x_i=x),
\end{align*}
where $p_\theta^X(x_i=x)$ is the neural network estimate for the probability that node $v_i$ in the clean graph $G$ is of type $x$.

\subsection{Adaptive Noise Schedule}

\begin{figure}[t]
    \centering
    \includegraphics[width=0.8\textwidth]{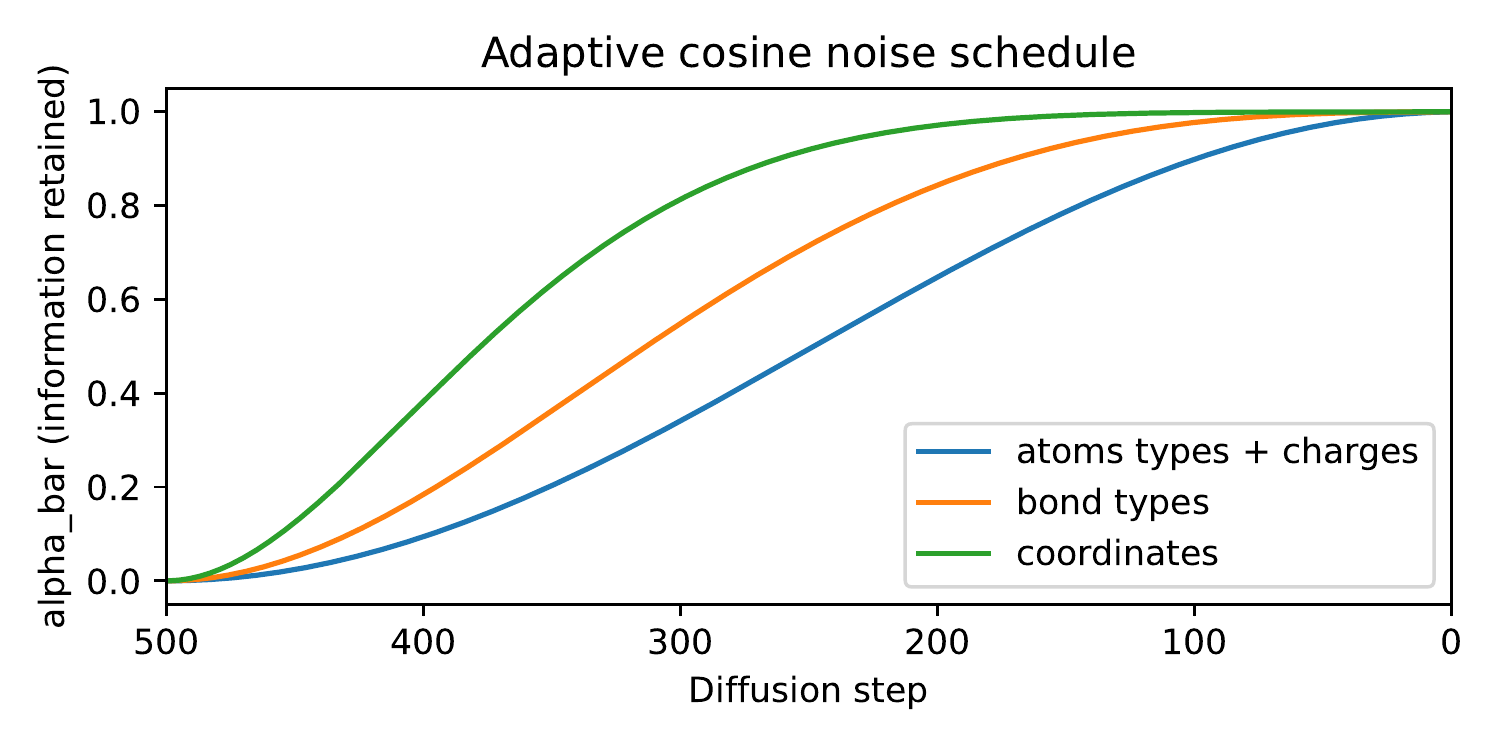}
    \vspace{-0.3cm}\caption{The noise schedule is tuned separately for each component. Atom coordinates and bond types are denoised earlier during sampling, while atom types and formal charges are updated later in the process. Experimentally, the adaptive schedule allows to obtain better 3D conformers and more stable molecules.} \label{fig:adaptive-schedule}
\end{figure}

Although the MiDi model corrupts the coordinates, atom types, bond types and formal charges simultaneously, these components do not play a symmetrical role. For instance, while the 2D connectivity structure can be predicted relatively well from the 3D conformation, the converse is not true as the conformation is not unique for a given structure. Similarly, the formal charges serve as an adjustable variable used to match the valency of each atom with its electronic structure, but they do not constitute a very fundamental property of the molecules.

Based on these observations, we propose an adaptation of the noise model in order to encourage the denoising network to first generate correctly the most important components, namely the atom coordinates and bond types, before moving on to predict the atom types and formal charges. To achieve this, we modify the noise schedule to vary according to the component. We modify the popular cosine schedule by adding an exponent $\nu$ that controls the rate at which the noise is added to the model:
$$
    \bar\alpha^t = \cos\left(\frac{\pi}{2} \frac{(t/T +s)^\nu}{1 + s}\right)^2,
$$
where the parameter $\nu$ can take the form of $\nu_r$, $\nu_x$, $\nu_y$ and $\nu_c$ for the atom coordinates, types, bond types, and charges, respectively. 
On the QM9 dataset, we use $\nu_r=2.5, ~ \nu_y=1.5, ~ \nu_x=\nu_c=1$, while GEOM-DRUGS uses $\nu_r=2$. The noise schedule used for QM9 is shown in Fig. \ref{fig:adaptive-schedule}.
This choice means that rough estimates for the atom coordinates and the bond types are first generated at inference, before the other components start to play a significant role. This aligns with previous work on 2D molecular graph generation which found that predicting the bond types before the atom types is beneficial \cite{madhawa2019graphnvp, vignac2021top}.

\subsection{Denoising Network}

\begin{figure}[h]  
    \centering
    \includegraphics[width=0.8\textwidth]{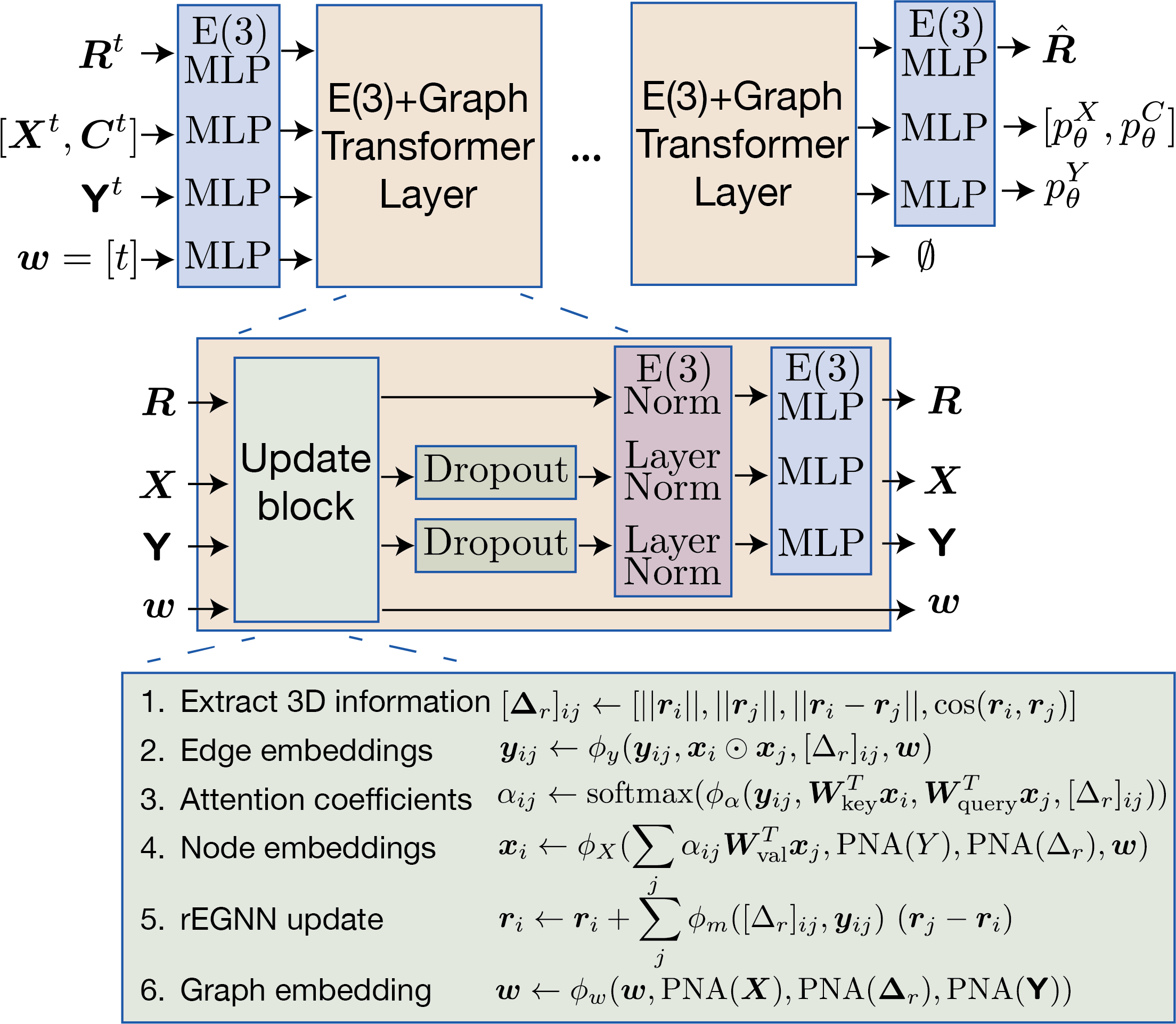}
    \caption{The denoising neural network of MiDi jointly predicts the 2D graph and 3D coordinates of the clean graph from a noisy input. It follows a graph Transformer architecture with layers tailored to maintain SE(3) equivariance. In the update block, each component is updated using the other features. While the graph-level features $\vw$ do not play a direct role in the final prediction, they serve as an effective means of storing and organizing pertinent information throughout the transformer layers.} \label{fig:midi-architecture}
\end{figure}

The denoising network takes a noisy graph as input and learns to predict the corresponding clean graph. It manipulates graph-level features $\vw$, node coordinates $\mR$, node features (atom types and formal charges, treated together in the matrix $\mX$), and edge features $\tY$. Coordinates are treated separately from the other node features in order to guarantee SE(3) equivariance. The neural network architecture is summarized in Figure \ref{fig:midi-architecture}. It consists of a Transformer architecture \cite{vaswani2017attention}, with a succession of self-attention module followed by normalization layers and feedforward networks. We give more details about the different blocks below.

\subsubsection{Relaxed Equivariant Graph Neural Networks (rEGNNs)}

In our proposed method, we leverage the effective yet affordable EGNN layers \cite{satorras2021egnn} for processing the coordinates. However, we enhance these layers by exploiting the fact that, when the data and the noise reside in the zero Center-Of-Mass subspace, it is not necessary for the neural network to be translation invariant. This can be interpreted as defining a canonical pose for the translation group, which is a valid way to achieve equivariance \cite{jaderberg2015spatial, kaba2022equivariance}.

Rather than simply relying on pairwise distances $||\vr_i - \vr_j||_2$, we can therefore use other rotation invariant descriptors such as $||\vr_i||_2$ or $\cos(\vr_i, \vr_j)$. We therefore propose the following \emph{relaxedEGNN} (rEGNN) layer:
\begin{align*}
    [\bm \Delta_r]_{ij} &= \operatorname{cat}(||\vr_i - \vr_j||_2, ||\vr_i||_2, ||\vr_j||_2, \cos(\vr_i, \vr_j)) \\
    \vr_i &\gets \vr_i + \sum_{j} \phi_m(\mX_i, \mX_j,  [\bm \Delta_r]_{ij}, \tY_{ij}) ~(\vr_j - \vr_i)
\end{align*}

Similar to EGNN layers, the rEGNN layer combines a rotation-invariant message function with a linear update in $\vr_j - \vr_i$, which guarantees rotation equivariance. Notably, the additional features $||\vr_i||_2$, $||\vr_j||_2$, and $\cos(\vr_i, \vr_j)$ are computed relative to the center-of-mass of the molecule, which is set to 0 by definition. In our experiments, we have observed that these features facilitate the generation of a higher proportion of connected molecules, thereby mitigating an issue previously observed with both 3D \cite{hoogeboom2022equivariant} and 2D-based denoising diffusion models \cite{vignac2022digress}.

\subsubsection{Update Block}

To improve the model's ability to process all features simultaneously, our new rEGNN layer is integrated into a larger update block that processes each component using all other ones.
The edge features are first updated using $\bm \Delta_r$, the node features, and the global features. The node features are updated using a self-attention mechanism, where the attention coefficients also use the edge features and $\bm \Delta_r$. After the attention heads have been flattened, the obtained values are modulated by the pooled edge features, $\bm \Delta_r$ and the global features. For pooling pairwise features ($\mY$ and $\bm \Delta_r$) into node representations, we use PNA layers \citep{corso2020principal}: $\operatorname{PNA}(\mY)_i = \mW^T \operatorname{cat}[\operatorname{mean}, \operatorname{min}, \operatorname{max}, \operatorname{std}]_j (\vy_{ij})$. The global features are updated by pooling all other features at the graph level. 
Finally, the coordinates are updated using a rEGNN update, where the message function takes as input $\bm \Delta_r$ and the updated edge features. Note that we do not use the normalization term of EGNN: our layers are integrated in a Transformer architecture as discussed next, and we empirically found SE(3) normalization layers to be more effective than the EGNN normalisation term at controlling the magnitude of the activations. 

\subsubsection{Integration into a Transformer Architecture}

Transformers have proved to be a very efficient way to stabilize the self-attention mechanism over many layers. We describe below the changes to the feed-forward neural network and normalization layers that are required to ensure SE(3)-equivariance.

Our feed-forward neural network processes each component using MLPs applied in parallel on each node and each edge. As the coordinates cannot be treated separately (it would break SE(3)-equivariance), we define
$$
\operatorname{PosMLP}(\mR) = \pcom (\MLP(||\mR||)  \frac{\mR}{||\mR|| + \delta}) \in \R^{n \times 3},
$$
where $||\mR|| \in \R^{n \times 1}$ contains the norm $||\vr_i||^2$ of each point, $\MLP(||\mR||) \in \R^{n \times 1}$ as well, $\delta$ is a small positive constant, and $\pcom$ is the projection of the coordinates on the linear subspace with center-of-mass at 0:
$$ \pcom(\mR)_i = \vr_i - \frac{1}{n} \sum_{i=1}^n \vr_i.$$

The choice of the normalization layer also depends on the problem symmetries: while batch normalization \cite{ioffe2015batch} is used in some graph transformer models \citep{dwivedi2020generalization}, this layer is not equivariant in contrast to  Set Normalization \cite{zhang2022set} or Layer Normalization \cite{ba2016layer}. For SE(3) equivariance, the normalization of \cite{liao2022equiformer} should be used. Applied to 3D coordinates, it writes
$$ \operatorname{E3Norm}(\mR) =  \gamma \frac{||\mR||}{\bar n + \delta} ~ \frac{\mR}{||\mR||} =  \gamma  \frac{\mR}{ \bar n + \delta} \quad \text{with} \quad \bar n = \sqrt{\frac{1}{n} \sum_{i=1}^n ||\vr_i||^2}, $$
with a learnable parameter $\gamma \in \R$ initialized at 1.

\subsection{Training Objective}

The denoising network of MiDi is trained to predict the clean molecule from a noisy input $G^t$, which is reflected in the choice of loss function used during model training. The estimation of the coordinates $\mR$ is a regression problem that can simply be solved with mean-squared error, whereas the prediction $p_\theta^X$ for the atom types, $p_\theta^C$ for the formal charges and $p_\theta^Y$ for the bond types corresponds to a classification problem which can be addressed through a cross-entropy loss ($\CE$ in the equations). Note that the network's position predictions result in pointwise estimates $\hat \mR$, while for the other terms, the prediction is a distribution over classes. The final loss is a weighted sum of these components:
\[
l(G, \hat p^G) = \lambda_r ||\hat \mR - \mR||^2 + \lambda_x \CE(\mX, p_\theta^X) + \lambda_c \CE(\mC, p_\theta^C) + \lambda_y \CE(\tY, p_\theta^Y)
\]

The $(\lambda_i)$ were initially chosen in order to balance the contribution of each term and cross-validated starting from this initial value. Our final experiments use $\lambda_r=3$, $\lambda_x=0.4$, $\lambda_c=1$, $\lambda_y=2$.

\section{Experiments}

\subsection{Settings}

We evaluate MiDi's performance on unconditional molecule generation tasks.
To the best of our knowledge, MiDi is the first method to generate both the graph structure and the conformer simultaneously, leaving no end-to-end differentiable method to compare to. We therefore compare MiDi to 3D models on top of which a bond predictor is applied. We consider two such predictors: either a simple lookup table, as used in \cite{hoogeboom2022equivariant}, or the optimization procedure of OpenBabel\footnote{\url{http://openbabel.org/wiki/Bond_Orders}} \cite{o2011open} used in other works such as \cite{igashov2022equivariant, schneuing2022structure}. The latter algorithm optimizes the bond orders of neighboring atoms in order to create a valid molecule, removing all control on the generated graphs. 
In terms of dataset comparison, EDM \cite{hoogeboom2022equivariant} was previously the only method that could scale up to the large GEOM-DRUGS dataset, so it is our only direct competitor in that case. For the QM9 dataset, we also compare MiDi's performance to that of the GSchNet method \citep{gebauer2019symmetry}, which employed the OpenBabel algorithm and achieved good results.

To facilitate comparison with previous methods, such as \citep{satorras2021en_flows} and \cite{hoogeboom2022equivariant}, we benchmark our models on the full molecular graphs that include explicit hydrogens atoms. However, we acknowledge that, for most practical applications, hydrogen atoms can be inferred from the heavy atoms in the structure, and thus can be removed. In fact, methods trained solely on heavy atoms usually perform better since they consider smaller graphs.

We measure validity using the success rate of RDKit sanitization over 10,000 molecules. Uniqueness is the proportion of valid molecules with different canonical SMILES. Atom and molecule stability are metrics proposed in \cite{satorras2021n} -- they are similar to validity but, in contrast to RDKit sanitization, they do not allow for adding implicit hydrogens to satisfy the valency constraints. Novelty is the proportion of unique canonical SMILES strings obtained that are not in the training set. Since all training molecules have a single connected component, we also measure the proportion of the generated molecules that are connected.

We also compare the histograms of several properties of the generated set with a test set. The atom and bond total variations (AtomTV and BondTV) measure the l1 distance between the marginal distribution of atom types and bond types, respectively, in the generated set and test set. The Wasserstein distance between valencies is a weighted sum over the valency distributions for each atom types: $\operatorname{ValencyW_1} = \sum_{x \in \text{atom types}} p(x) \mathcal W_1(\hat  D_\text{val}(x),  D_\text{val}(x))$, where $p^X(x)$ is the marginal distribution of atom types in the training set, $\hat D_\text{val}(x)$ is the marginal distribution of valencies for atoms of type $x$ in the generated set, $D_\text{val}(x)$ the same distribution in the test set. Here, the Wasserstein distance between histograms is used rather than total variation, as it allows to better respect the structure of ordinal data.

In previous methods, graph-based metrics were predominantly used. However, in our approach, we also introduce 3D metrics based on histograms of bond lengths and bond angles. This allows us to evaluate the efficacy of our approach not only in terms of the graph structure but also in generating accurate conformers. To this end, we report a weighted sum of the distance between bond lengths for each bond type:  
$$\operatorname{BondLenghtsW_1} = \sum_{y \in \text{bond types}} p(y) \mathcal W_1(\hat  D_\text{dist}(y),  D_\text{dist}(y)),$$
where $p^Y(y)$ is the proportion of bonds of type $y$ in the training set, $\hat D_\text{dist}(y)$ is the generated distribution of bond lengths for bond of type $y$, and $D_\text{dist}(y)$ is the same distribution computed over the test set. The output is value in Angstrom.

Finally, $\operatorname{BondAnglesW_1}$ (in degrees) compares the distribution of bond angles (in degrees) for each atom type. We compute a weighted sum of these values using the proportion of each atom type in the dataset. This calculation is restricted to atoms with two or more neighbors to ensure that angles can be defined:
$$\operatorname{BondAnglesW_1}(\text{generated}, \text{target}) = \sum_{x \in \text{atom types}} \tilde p(x) \mathcal W_1(\hat  D_\text{angles}(x),  D_\text{angles}(x)),
$$
where $\tilde p^X(x)$ denotes the proportion of atoms of type $x$ in the training set, restricted to atoms with two neighbors or more, and $D_\text{angles}(x)$ is the distribution of geometric angles of the form $\angle(\vr_k - \vr_i, \vr_j - \vr_i)$, where $i$ is an atom of type $x$, and $k$ and $j$ are neighbors of $i$. The reported metrics are mean and 95\% confidence intervals on 5 different samplings from the same checkpoint. 

\subsection{QM9}

\begin{table}[t]
    \centering
        \caption{Unconditional generation on QM9 with explicit hydrogens with uniform and adaptive noise schedules. While MiDi outperforms the base EDM model on graph-based metrics, the Open Babel optimization procedure is very effective on this simple dataset, as the structures are simple enough for the bonds to be determined unambiguously from the conformation.}
    \begin{tabular}{l l l l l l l} \toprule
    \textit{Metrics ($\uparrow$)} & Mol stable & At stable & Validity & Uniqueness &  Novelty & Connected   \\  
    Data                & $98.7$     &  $99.8$   & $98.9$ & $99.9$ &  -- & $100.0$ \\\midrule
    GSchNet    & $92.0$     &  $98.7$   & $98.1$ & $94.5$ & $\bm{80.5}$ & $97.1$  \\
    EDM                 & $90.7$     &  $99.2$   & $91.7$ & $\bm{98.5}$ & $75.9$ & $99.3$   \\
    EDM + OBabel        & $\bm{97.9}$ & $\bm{99.8}$ & $\bm{99.0}$ & $\bm{98.5}$ & $77.8$ & $99.7$   \\
    MiDi  (uniform) & $96.1\spm{.2}$ &  $\bm{99.7}\spm{.0}$ & $96.6\spm{.2}$ & $97.6\spm{.1}$ & $64.9\spm{.5}$ & $99.8\spm{.0}$ \\
    MiDi (adaptive) & $\bm{97.5}\spm{.1}$ & $\bm{99.8}\spm{.0}$ & $97.9\spm{.1}$ & $97.6\spm{.1}$ & $67.5\spm{.3}$ & $\bm{99.9}\spm{.0}$\\ \midrule
    \textit{Metrics ($\downarrow)$} & Valency(e-2) & Atom(e-2) & Bond(e-2)   & Angles (°) &\multicolumn{2}{l}{Bond Lengths (e-2 Å)}  \\
    Data & $0.1$  & $0.3$ & $\sim 0$ &  $0.12$ & \multicolumn{2}{l}{$\sim0$}\\ \midrule
    GSchNet & $4.9$ & $4.2$ & $1.1$  &  $1.68$ & \multicolumn{2}{l}{$0.5$} \\
    EDM & $1.1$ &$2.1$ & $0.2$    & $\bm{0.44}$ & \multicolumn{2}{l} {$\bm{0.1}$}\\
    EDM + OBabel & $1.1$ & $2.1$ & $\bm{0.1}$ & $\bm{0.44}$ & \multicolumn{2}{l}{$\bm{0.1}$}\\
    MiDi (uniform) & $0.4\spm{.0}$&  $0.9\spm{.0}$ & $\bm{0.1}\spm{0.0}$  & $0.67\spm{.02}$ &\multicolumn{2}{l}{$1.6\spm{.7}$}\\
    MiDi (adaptive) & $\bm{0.3}\spm{.0}$ & $\bm{0.3}\spm{.1}$ & $\bm{0.0}\spm{.0}$  & $0.62\spm{.02}$ & \multicolumn{2}{l}{$0.3\spm{.1}$}\\
    \bottomrule
    \end{tabular}
    \label{tab:qm9-h}
\end{table}

We first evaluate our model on the standard QM9 dataset \cite{wu2018moleculenet} containing molecules with up to 9 heavy atoms. We split the dataset into 100k molecules for training, 20k for validation, and 13k for testing. Results are presented in Table \ref{tab:qm9-h}. The \textit{data} line represents the results of the training set compared with the test set, while  the other entries compare the generated molecules to the test molecules. As we observe in Table \ref{tab:qm9-h}, predicting the bonds only from the interatomic distances and atom types has limited performance. Therefore, MiDi outperforms EDM on 2D metrics, while obtaining similar 3D metrics for the generated conformers. It is worth noting that our list of allowed bonds is not identical to that used in \cite{satorras2021n, hoogeboom2022equivariant}, which may explain why our results for EDM \cite{hoogeboom2022equivariant} do not match those of the original paper perfectly. Nonetheless, the optimization algorithm of OpenBabel performs very well on this dataset of simple molecules. As QM9 contains molecules with only up to 9 atoms, the molecular conformations are easy to understand and the bonds can be determined easily.

\subsection{GEOM-DRUGS}

\begin{table}[t]
    \centering
        \caption{Unconditional generation on GEOM-Drugs with explicit hydrogens. EDM was previously the only method that scaled to this dataset. On this complex dataset, the benefits of an integrated models are very clear, as MiDi significantly outperforms Open Babel on most metrics. 95\% confidence intervals are reported on five samplings of the same checkpoint.}
    \begin{tabular}{l l l l l l l} \toprule
    \textit{Metrics ($\uparrow$)} & Mol stable & At stable & Validity & Uniqueness &  Novelty & Connected   \\  
    Data                & $99.9$     &  $99.9$   & $99.8$ & $100.0$ &  -- & $100.0$ \\\midrule
    EDM                 & $5.5$     &  $92.9$   & $\bm{97.5}$ & $99.9$ & $\bm{100.0}$ & $35.6$   \\
    EDM + OBabel        & $40.3$ & $97.8$ & $87.8$ & $99.9$ & $\bm{100.0}$ & $41.4$   \\
    MiDi  (uniform) & $89.9\spm{.2}$ & $99.7\spm{.0}$ & $74.5\spm{.2}$ & $\bm{100.0}\spm{.0}$ & $\bm{100.0}\spm{.0}$ & $\bm{90.5}\spm{.2}$ \\
    MiDi (adaptive) & $\bm{91.6}\spm{.2}$ & $\bm{99.8}\spm{.0}$ & $77.8\spm{.2}$ & $\bm{100.0}\spm{.0}$ & $\bm{100.0}\spm{.0}$ & $90.0\spm{.3}$\\ 
    \midrule
    \textit{Metrics ($\downarrow)$} & Valency(e-2) & Atom(e-2) & Bond(e-2)   & Angles (°) &\multicolumn{2}{l}{Bond Lengths (e-2 Å)}  \\
    Data & $0.1$  & $0.1$ & $2.5$ &  $0.05$ & \multicolumn{2}{l}{$\sim0$}\\ \midrule
    EDM & $11.2$ & $21.2$ & $4.9$    & $6.23$ & \multicolumn{2}{l} {$\bm{0.2}$}\\
    EDM + OBabel & $28.5$ & $21.2$ & $4.8$ & $6.42$ & \multicolumn{2}{l}{$\bm{0.2}$}\\
    MiDi (uniform) & $2.9\spm{.0}$ & $3.9\spm{.1}$ & $\bm{2.4}\spm{.0}$  & $1.43\spm{.002}$ &\multicolumn{2}{l}{$1.1\spm{.2}$}\\
    MiDi (adaptive) & $\bm{0.8}\spm{.1}$ & $\bm{3.8}\spm{.1}$ & $\bm{2.4}\spm{.0}$ & $\bm{1.07}\spm{.02}$ & \multicolumn{2}{l}{$\bm{0.2}\spm{.1}$}\\
    \bottomrule
    \end{tabular}
    \label{tab:geom-with-h}
\end{table}

We then assess our model on the much larger GEOM-DRUGS dataset \cite{axelrod2020geom} which comprises 430,000 drug-sized molecules with an average of 44 atoms and up to 181 atoms. As this dataset features drug-like compounds, it is therefore better suited for downstream applications than QM9. We split the dataset into 80\% for training, 10\% for validation, and 10\% for testing. For each molecule, we extract the 5 lowest energy conformations to build the dataset. Results are presented in Table \ref{tab:geom-with-h}. On this large dataset, we did not train the adaptive version of MiDi from scratch, but instead fine-tuned it using a checkpoint of MiDi with uniform noise schedule.

As this dataset contains molecules that are much more complex than those in QM9, the bonds in the molecules cannot be determined solely from pairwise distances. This explains why EDM, which performs relatively well on 3D-based metrics, produces very few valid and stable molecules. Furthermore, many structures in this dataset are too complex for the Open Babel algorithm. While the latter achieves good atom stability, there is at least one invalid atom in most molecules, leading to low molecular stability. The advantages of an end-to-end model that generates both a graph structure and its conformation are evident on this dataset: MiDi not only generates better molecular graphs, but also predicts 3D conformers with more realistic bond angles.

\section{Conclusions}

We propose MiDi, a novel denoising diffusion model that jointly generates a molecular graph and a corresponding conformation for this molecule. Our model combines Gaussian and discrete diffusion in order to define a noise model that is best suited to each component. The noise schedule is further adapted to the different components, with the network initially generating a rough estimate of the conformation and the graph structure, before tuning the atom types and charges. A graph transformer network is trained to denoise this model, that features novel rEGNN layers. While rEGNN layers manipulate features that are translation invariant, they still result in a SE(3) equivariant network when the input molecules are centered. 
On unconditional generation tasks with complex datasets, MiDi clearly outperforms prior 3D molecule generation methods that predict bonds from the conformation using predefined rules. While our model is currently evaluated on unconditional generation tasks, we believe that the end-to-end training of the graph structure and the conformation can offer even greater benefits for other downstream tasks such as pocket-conditioned generation.

\section{Acknowledgements}
This work was partially supported by The Alan Turing Institute. Clément Vignac thanks the Swiss Data Science Center for supporting him through the PhD fellowship program (grant P18-11).

\bibliographystyle{splncs04}
\bibliography{biblio}

\appendix

\section{Additional Results} \label{appendix:additional-results}

\subsection{QM9 with Implicit Hydrogens}

The results are presented in Table \ref{tab:qm9-no-h} for 5 samplings of the same checkpoint. While all methods overall achieve good results on this simple dataset, we observe that the lookup table of EDM sometimes fails to predict the bond type, resulting in much more invalid molecules than our model. Interestingly, the adaptive noise schedule that allowed for important improvements on the GEOM-DRUG dataset is not effective on this simpler dataset, and the uniform schedule seems to perform better. The reasons for this phenomenon. We finally observe that the bond length predictions are overall good for all methods, but that MiDi is not as precise as EDM, which can be explained by the fact that EDM uses both learning rate decay and an exponential moving average.

\begin{table}[t]
    \centering
        \caption{Unconditional generation on QM9 with implicit hydrogens. On this simple dataset, all methods achieve very good results, although the lookup table of EDM sometimes fail to generate correct edge types, resulting in invalid molecules.}
    \begin{tabular}{l l l l l} \toprule
    Metric ($\uparrow$) & Validity & Uniqueness & Novelty & Connected  \\
        Data & $99.5$ & $99.9$ & -- &  100 \\ \midrule 
    EDM   &  $96.8$ & $\bm{96.6}$ & $45.5$ & $99.9$ \\
    EDM + OBabel & $\bm{100.0}$ & $96.1$ & $45.4$ & $99.9$ \\
    MiDi (uniform) & $99.5\spm{.1}$ & $95.8\spm{.2}$ & $\bm{49.2}\spm{0.4}$& $\bm{100}\spm{.0}$\\
    MiDi (adaptive) & $99.7\spm{.0}$ & $93.9\spm{.2}$ &  $44.7\spm{.5}$  & $\bm{100}\spm{.0}$ \\ \midrule
    Metric ($\downarrow$) & Valency(e-2) & Atom(e-2) & Bond(e-2) & Bond Lengths(e-2 Å) \\
    Data &  $0.6$ & $0.1$ & $0.1$ & $0.3$ \\\midrule
    EDM & $4.3$ & $\bm{2.9}$ & $0.9$ & $\bm{0.2}$\\
    EDM + OBabel &  $3.8$ & $\bm{2.9}$ & $\bm{0.3}$ & $\bm{0.2}$ \\
    MiDi (uniform) & $\bm{2.3}\spm{1.7}$ & $3.1\spm{3.7}$ & $\bm{0.4}\spm{.1}$ &  $0.8\spm{.3}$\\
    Midi (adaptive) & $5.9\spm{.2}$ & $11.7\spm{0.2}$ & $1.1\spm{.0}$ &  $1.6\spm{.7}$ \\
    \bottomrule
    \end{tabular}
    \label{tab:qm9-no-h}
\end{table}

\subsection{GEOM-DRUGS with implicit hydrogens}
\begin{table}[t]
    \centering
        \caption{Unconditional generation on GEOM-DRUGS with implicit hydrogens.}
    \begin{tabular}{l l l l l} \toprule
    Metric ($\uparrow$) & Validity & Uniqueness & Novelty & Connected  \\
        Data & $100.0$ & $100.0$ & -- &  $99.8$ \\ \midrule 
    MiDi (uniform) & $93.2\spm{.1}$ & $\bm{100.0}\spm{.0}$ & $\bm{100.0}\spm{.0}$& $98.0\spm{.1}$\\
    MiDi (adaptive) & $\bm{96.7}\spm{.1}$ & $\bm{100.0}\spm{.0}$ & $\bm{100.0}\spm{.0}$  & $\bm{98.7}\spm{.1}$ \\ \midrule
    Metric ($\downarrow$) & Valency(e-2) & Atom(e-2) & Bond(e-2) & Bond Lengths(e-2 Å) \\
    Data &  $\sim0$ & $\sim0$ & $7.9$ & $\sim0$ \\\midrule
    MiDi (uniform) & $13.6\spm{.1}$ & $9.5\spm{.3}$ & $8.8\spm{.0}$ &  $1.6\spm{.2}$\\
    Midi (adaptive) & $\bm{5.3}\spm{.1}$ & $\bm{6.3}\spm{0.1}$ & $\bm{8.0}\spm{.0}$ & $\bm{0.6}\spm{.3}$ \\
    \bottomrule
    \end{tabular}
    \label{tab:geom-no-h}
\end{table}

\section{Samples from our model}
\begin{figure}[h]
    \centering
    \includegraphics[width=\textwidth]{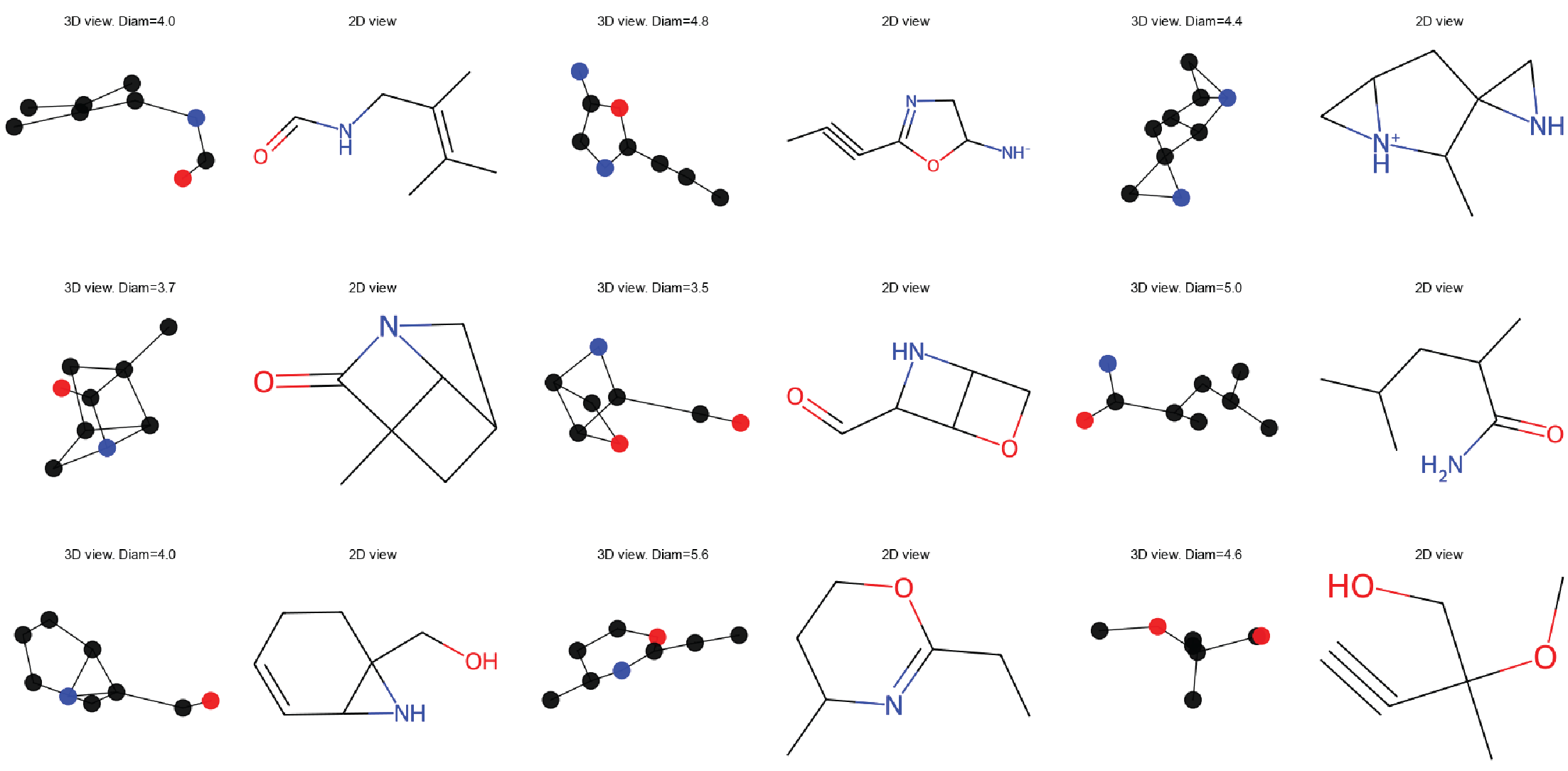}
    \caption{Non-curated samples on QM9 with implicit hydrogens.}
    \label{fig:samples-qm9-no-h}
\end{figure}
\begin{figure}[h]
    \centering
    \includegraphics[width=\textwidth]{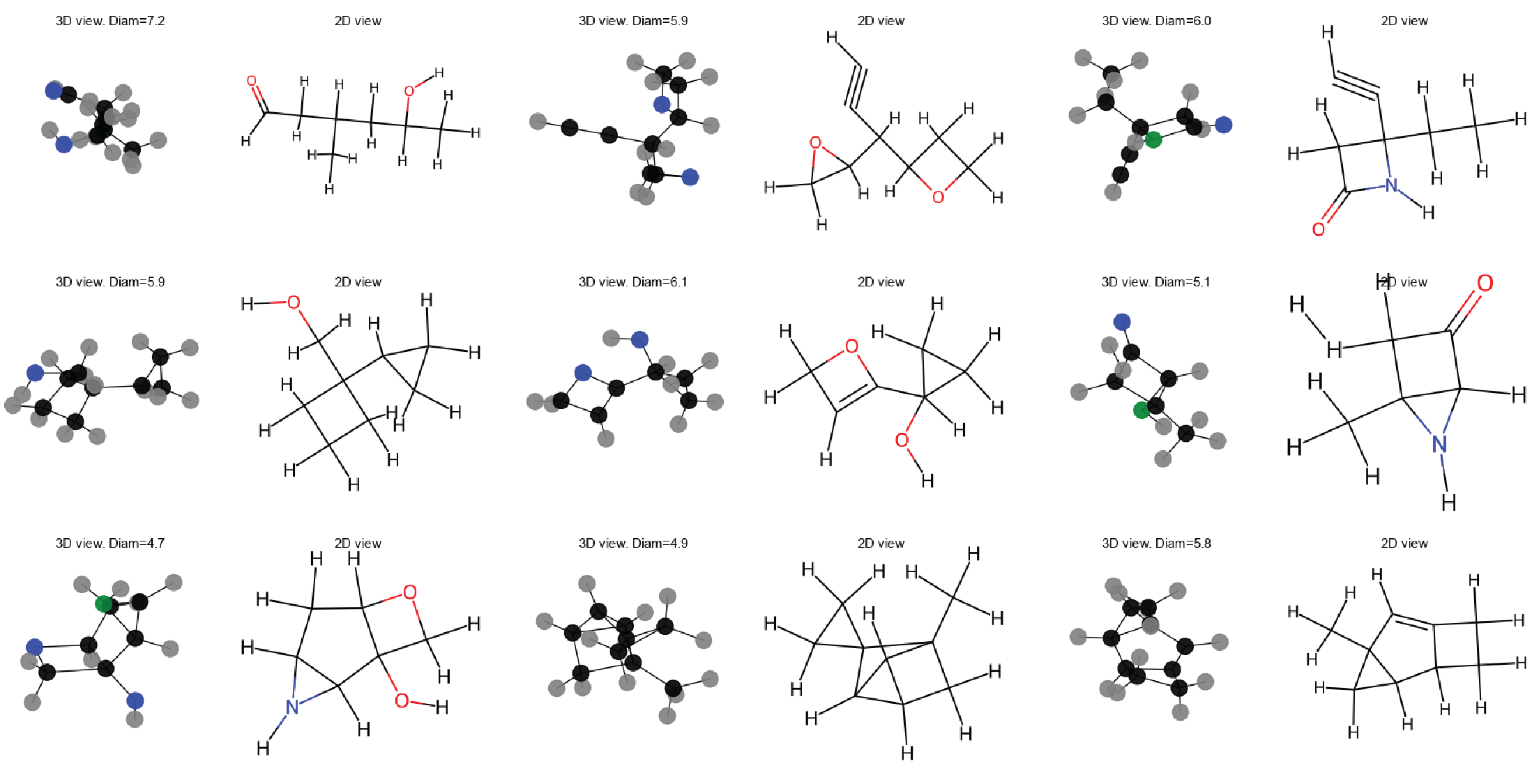}
    \caption{Non-curated samples on QM9 with explicit hydrogens.}
    \label{fig:samples-qm9-h}
\end{figure}
\begin{figure}[h]
    \centering
    \includegraphics[width=\textwidth]{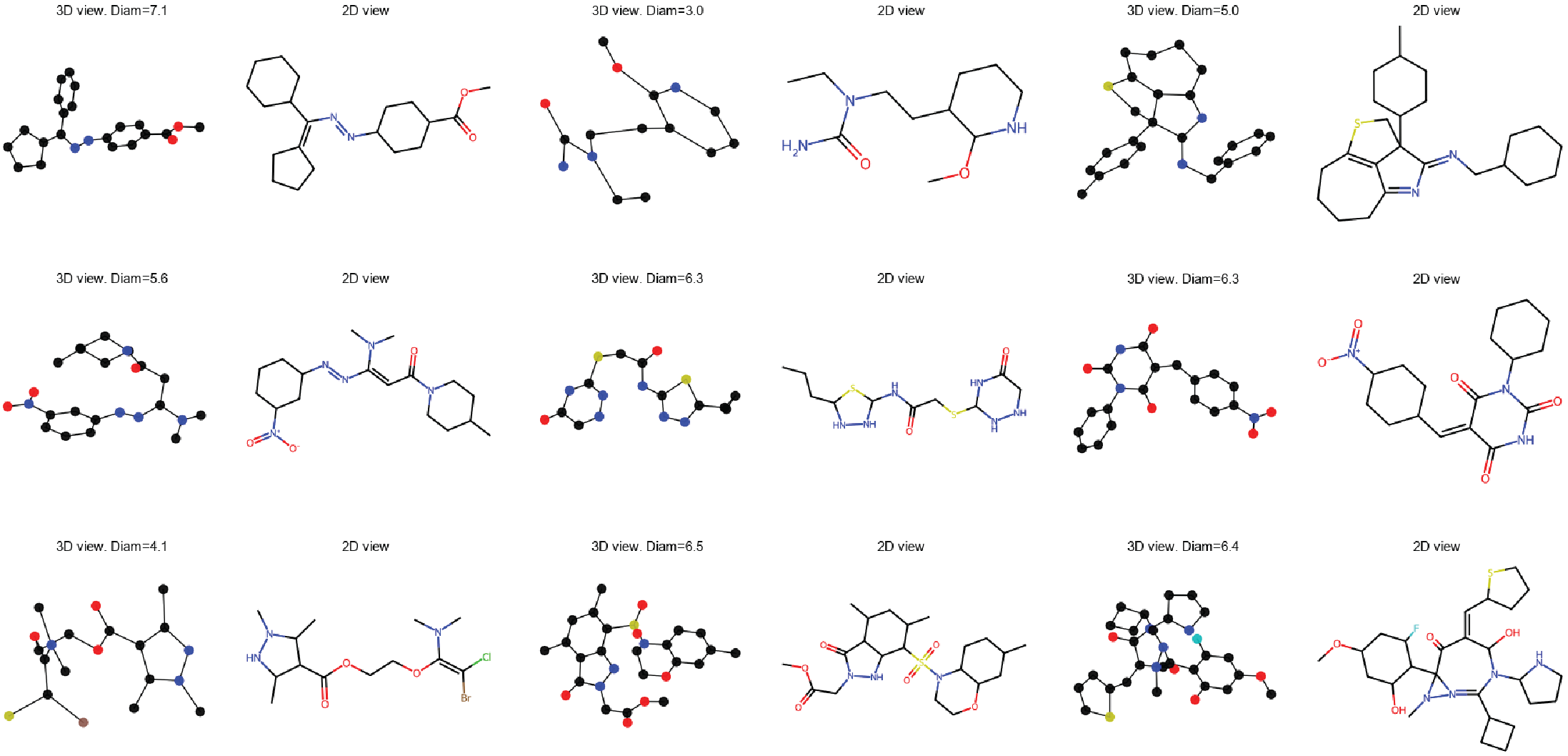}
    \caption{Non-curated samples on GEOM-drugs with implicit hydrogens.}
    \label{fig:samples-geom-no-h}
\end{figure}
\begin{figure}[h]
    \centering
    \includegraphics[width=\textwidth]{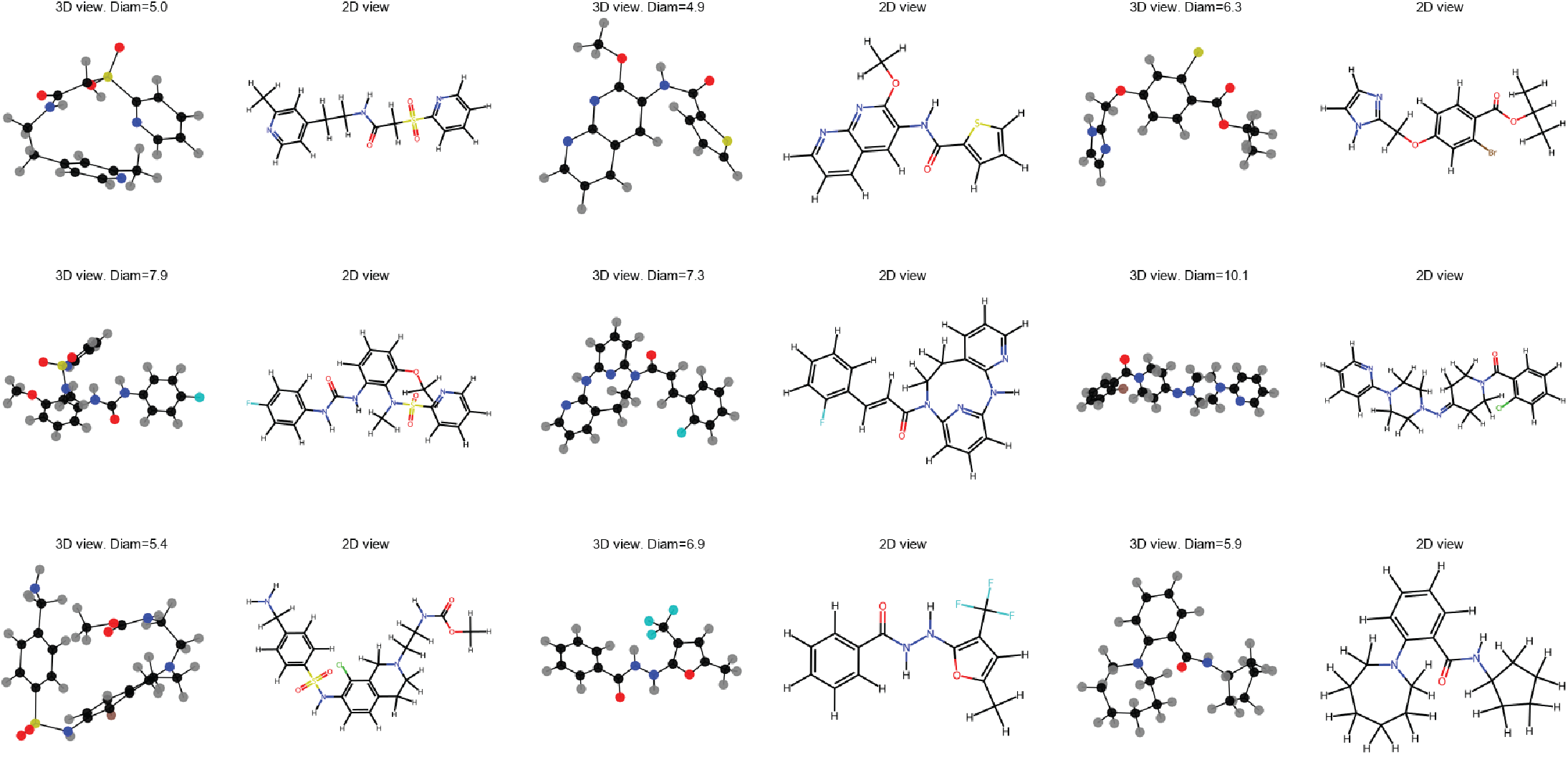}
    \caption{Non-curated samples on QEOM-drugs with explicit hydrogens.}
    \label{fig:samples-geom-h}
\end{figure}

\end{document}